\documentclass[a4paper,twoside]{article}

\usepackage{epsfig}
\usepackage{subfigure}
\usepackage{calc}
\usepackage{amssymb}
\usepackage{amstext}
\usepackage{amsmath}
\usepackage{amsthm}
\usepackage{multicol}
\usepackage{pslatex}
\usepackage{apalike}
\usepackage{enumitem}
\usepackage{SCITEPRESS}     

\begin{document}

\title{A Convolutional Neural Network for Language-Agnostic Source Code Summarization}

\author{\authorname{Jessica Moore\sup{1}\orcidAuthor{0000-0003-2076-3796}, Ben Gelman\sup{1}\orcidAuthor{0000-0002-1184-4116} and David Slater\sup{1}\orcidAuthor{0000-0001-5639-0253}}
\affiliation{\sup{1}Machine Learning Group, Two Six Labs, Arlington, Virginia, United States}
\email{\{jessica.moore, ben.gelman, david.slater\}@twosixlabs.com}}

\keywords{Natural Language Processing, Deep Learning, Source Code,  Crowdsourcing, Automatic Documentation}

\abstract{Descriptive comments play a crucial role in the software engineering process. They decrease development time, enable better bug detection, and facilitate the reuse of previously written code. However, comments are commonly the last of a software developer's priorities and are thus either insufficient or missing entirely. Automatic source code summarization may therefore have the ability to significantly improve the software development process. We introduce a novel encoder-decoder model that summarizes source code, effectively writing a comment to describe the code's functionality. We make two primary innovations beyond current source code summarization models. First, our encoder is fully language-agnostic and requires no complex input preprocessing. Second, our decoder has an open vocabulary, enabling it to predict any word, even ones not seen in training. We demonstrate results comparable to state-of-the-art methods on a single-language data set and provide the first results on a data set consisting of multiple programming languages.}

\onecolumn \maketitle \normalsize \vfill

\section{\uppercase{Introduction}}
\label{sec:introduction}

\noindent Studies of software development patterns suggest that software engineers spend a significant amount of their productive time on program comprehension tasks, such as reading documentation or trying to understand a colleague's code \cite{xia2017measuring,minelli2015know,ko2006exploratory}. Experiments, developer interviews, and studies of open-source systems all confirm that accurate comments are critical to effective software development, maintenance, and evolution \cite{xia2017measuring,wong2013autocomment,ibrahim2012relationship,de2005study}. Comments enable programmers to understand code more rapidly, prevent them from duplicating existing functionality, and aid them in fixing (or preventing) bugs. Descriptive comments even enable programmers to conduct natural language searches for code of interest. Unfortunately, many codebases suffer from a lack of thorough documentation \cite{steinmacher2014hard,parnas2011precise,briand2003software}. Retroactive manual documentation is increasingly expensive and infeasible, due to the growing volume of code \cite{deshpande2008total}. Automatic summarization of source code, therefore, holds the potential to significantly improve the software development life-cycle by filling in these gaps -- adding descriptive comments where the developers themselves did not do so. We address the question of whether neural machine translation methods can be used to automatically write comments for arbitrary source code. 

Current state-of-the-art source code summarization models take a language-specific approach, using lexemes (i.e., lexer output) or abstract syntax trees (ASTs) as model input \cite{hu2018deep,iyer2016summarizing}. Deploying these algorithms broadly would require many language-specific models, each with its own parser, training set, and hyperparameters. Writing consistent parsers, collecting data sets, training/tuning the models, and deploying them all simultaneously would be exceedingly burdensome for any reasonably sized set of languages. 

Additionally, current source code summarization models employ ``closed'' vocabularies, i.e., all of the words that can be predicted are known in advance. However, source code comments often contain words that are made-up (e.g., agglutinative method names, such as ``isValid'') or highly project-specific (e.g., ``playerID''). Models with closed vocabularies face a trade-off: they can either employ a relatively small vocabulary that will severely limit descriptiveness or they can employ a relatively large vocabulary (and thus a high degree of model complexity) and risk overfitting.

We introduce a novel encoder-decoder model designed to overcome both of these limitations. Specifically, we construct a deep convolutional encoder that directly ingests source code as a sequence of characters. This source-code-as-text approach ensures that the model is fully language-agnostic and requires no complex input preprocessing (e.g., parsing, lexing, etc.). Additionally, we construct a novel, ``open-vocabulary'' decoder that can predict words, subwords, and single characters. By combining those word components, the decoder is capable of generating arbitrary words without making use of an extensive vocabulary. And, because it incorporates words, subwords, and single characters, our model learns word meaning in a more substantive manner than other open-vocabulary models.


Our primary contributions are:
\begin{itemize}
	\item Introducing an encoder model capable of ingesting arbitrary source code (multiple languages, incorrect syntax, etc.).
    \item Introducing a novel vocabulary creation method that allows the model to effectively overcome the long-tailed nature of terms in source code comments.
    \item Demonstrating state-of-the-art summarization results on a single-language data set and providing the first summarization results on a multi-language data set.
\end{itemize}

The remainder of the paper is organized as follows: In Section 2, we discuss our approach to the task; in Section 3, we outline the experiments that we conduct; in Section 4, we present the results of these experiments; in Section 5, we review related research and compare it to our own; and, in Section 6, we draw conclusions and propose future work.

\section{\uppercase{Approach}}

\begin{figure}[h]
   \centering
   \includegraphics[scale=.20]{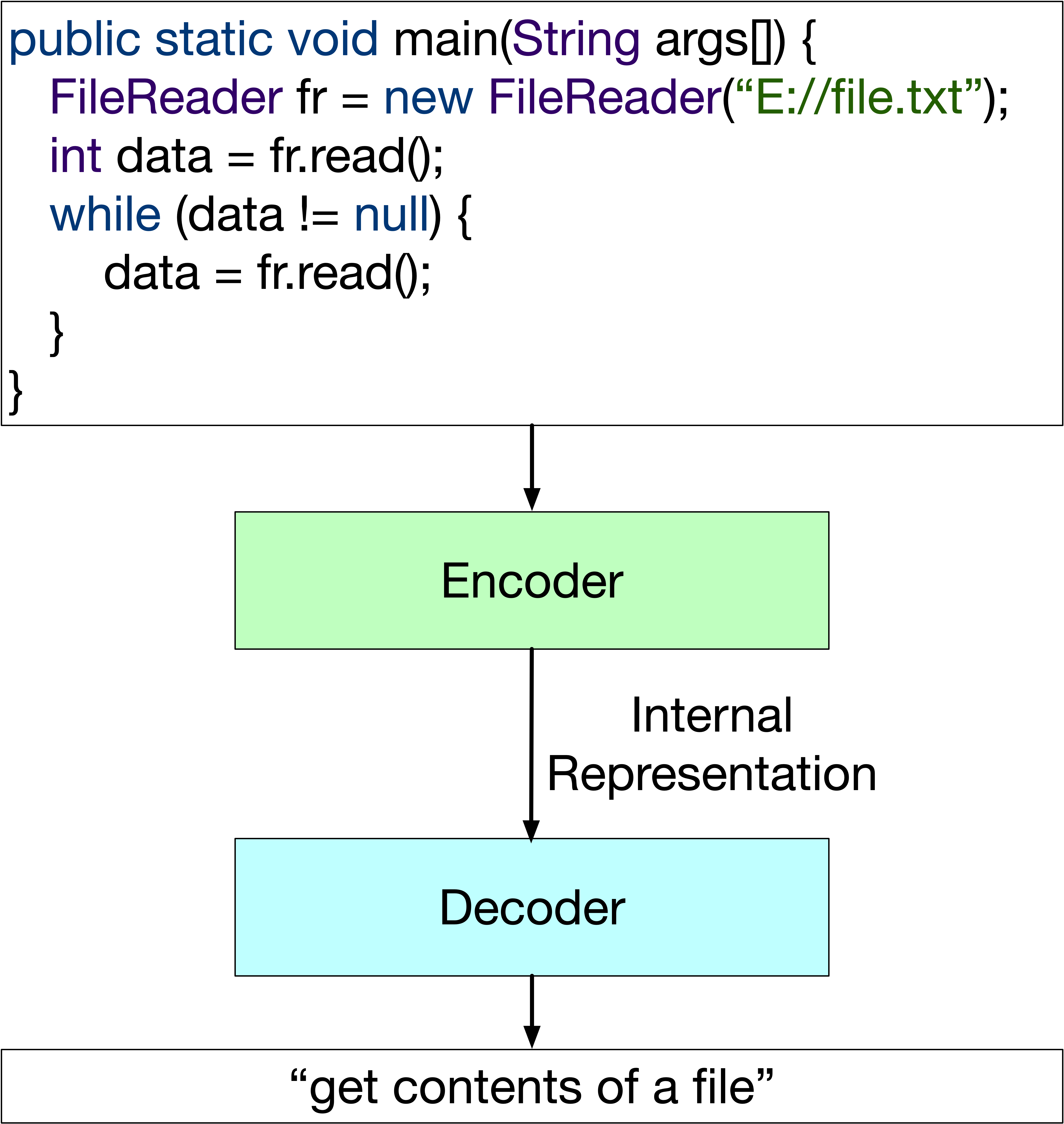}
    \caption{High-level model architecture. The model is composed of an encoder (green), which ingests the code and generates an internal representation of it, and a decoder (blue), which uses the internal representation of the code to produce an appropriate comment. Here, "get contents of a file" is the true (target) label.}
    \label{fig:encoder_decoder_architecture}
\end{figure}

\noindent We develop a deep neural network to generate natural language summaries of source code.  Figure \ref{fig:encoder_decoder_architecture} contains a demonstrative example (which we will use going forward) and depicts a high-level sketch of our model. The model has two primary components: an encoder, which reads in code and generates an internal representation of it, and a decoder, which uses the internal representation to generate a descriptive summary.

    \subsection{Convolutional Encoder}
 
\noindent In order to ingest code of arbitrary language and complexity, we employ a character-level approach, viewing the code as a sequence of characters, instead of a sequence of tokens. Character-level models have proven effective in both the natural language and source code domains \cite{kim2016character,gelman2018language}. 

\begin{figure}[h]
   \centering
   \includegraphics[scale=.20]{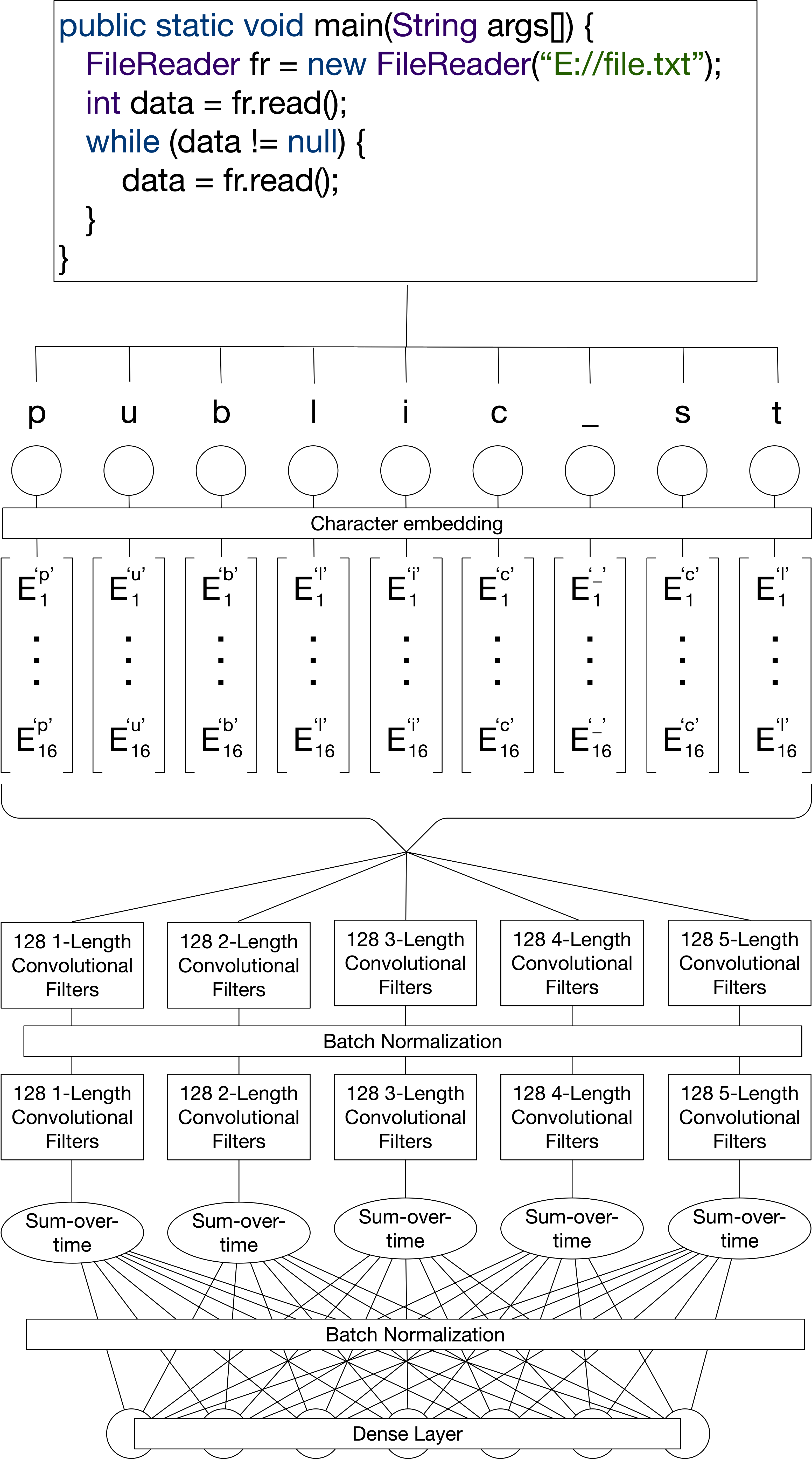}
    \caption{Encoder architecture. The character-level ingestion of code ensures complete language- and syntax-agnosticism. Subsequent convolutional layers allow the model to learn the significance of different character combinations.}
    \label{fig:encoder_architecture}
\end{figure}

Figure \ref{fig:encoder_architecture} depicts the encoding process. Each unique character in the input is mapped to a fixed-dimensional vector (``character embedding''). All instances of the letter ``c'', for example, are mapped to the same vector, regardless of where they appear in the input code. These character embeddings are learned in conjunction with the model weights. After each character is mapped to an embedded space, two layers of 1-dimensional convolutions and a single layer of sum-over-time pooling are performed on the sequence. The X-length convolutional filters operate on X consecutive character embeddings at a time. For example, in Figure \ref{fig:encoder_architecture}, the 3-length convolutional filters in the first layer would operate on the character embeddings for the sequences ``pub'', ``ubl'', etc. The convolutional filters learn to usefully combine the character embeddings to obtain information about character sequences of interest. In effect, the model is capable of learning much the same information about a token's meaning as would usually be captured by a token embedding. However, a character-level approach generalizes significantly better, because tokens with related meaning often contain similar character patterns (e.g., ``FileReader'' and ``FileWriter''). The sum-over-time pooling layer enables the model to create a fixed-length vector from a variable-length input, allowing the model to ingest code of arbitrary length. After the pooling layer, a final dense layer is applied. The resultant vector, sometimes referred to as a ``thought vector,'' is the model's internal representation of the code. By combining convolutional results via the pooling and dense layers, the model will be able to internally represent entire concepts in the input code, such as ``public''.

Previous work on source code summarization has used token-level embeddings \cite{iyer2016summarizing,hu2018deep}. However, in any realistic corpus of source code, many tokens do not appear frequently enough for a meaningful embedding to be learned. For example, a variable name may appear in only a single piece of code.
Recognizing this, previous works have used generic identifiers (e.g., ``column0", ``SimpleName", or ``Unknown") in place of low-occurrence tokens \cite{iyer2016summarizing,hu2018deep}. However, this practice prevents the model from capitalizing on some of the semantic information available in the input source code. Our character-based approach allows the model access to that semantic information without requiring it to learn an unreasonable number of token embeddings. Perhaps more significantly, our character-level approach is critical to the model's language agnosticism. Tokenization is necessarily language-dependent, because the meaning of punctuation is language-dependent. Since our model focuses on characters, instead of higher-level constructs (e.g., tokens), it can ingest code of any language, without placing constraints on the code's syntactic correctness.
    
    \subsection{LSTM Decoder}
    
\noindent Our decoder translates the ``thought vector'' into natural language. As with most machine translation models, the decoder is composed of long short-term memory (LSTM) units. To transmit information from the encoder to the decoder, the initial hidden state of the LSTM is set equal to the thought vector from the encoder.  The LSTM generates a sequence of predictions, which are combined to form the predicted comment. Per Section \ref{sec:mandtdetails}, we adjust the number of LSTM layers in the decoder based on the data set.

Unlike those of previous machine translation models, our model's vocabulary consists of words, subwords (e.g., ``ing''), single characters, punctuation, and special tokens (e.g., ``START''). See Section \ref{sec:vocab} for details on the vocabulary selection process. At each step in the decoding process, the model produces a probability distribution over the elements in the vocabulary. We train the model using cross-entropy loss across the entire sequence. 

\begin{figure}[h]
   \centering
   \includegraphics[scale=.35]{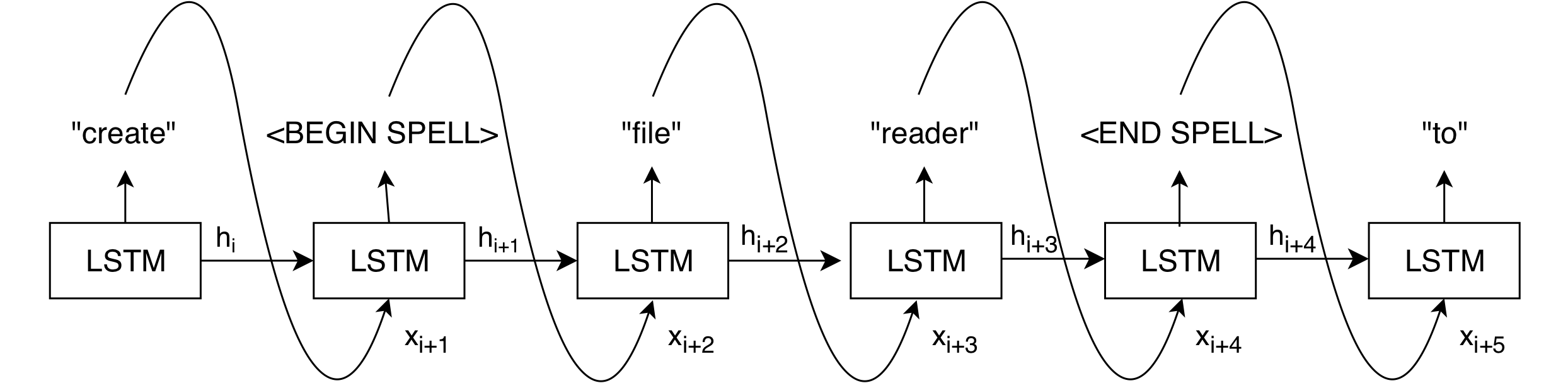}
    \caption{Decoder, in test mode, predicting the phrase ``create filereader to.'' Because the model is able to spell out any word, our method entirely avoids unknown word tokens, which are conventionally used in place of words that are outside of a model's vocabulary.}
    \label{fig:decoder_architecture2}
\end{figure}

To create the final predicted sentence, we select the most-probable element at each step. Sequential predictions can be combined to form a single word. In order to determine when the predictions should be combined (e.g., identifying when ``I'' should be a component of ``playerID'' instead of its own word), we add two special tokens to the model's vocabulary: ``BEGIN SPELL'' and ``END SPELL''. Outputs predicted between these two tokens are combined into one word. This process is depicted in Figure~\ref{fig:decoder_architecture2}. For simplicity, we show a single-layer LSTM.

To train the model, we tokenize the target comment so that it is made up of a series of words (determined by space-separation) and punctuation marks. For any word in the target comment that is not in the vocabulary, we greedily divide it so that it is composed of elements in the vocabulary. E.g., if ``filenotfound'' is not in the vocabulary but ``file'', ``not'', and ``found'' are, the compound term will be divided accordingly.

\begin{figure}[h]
   \centering
   \includegraphics[scale=.50]{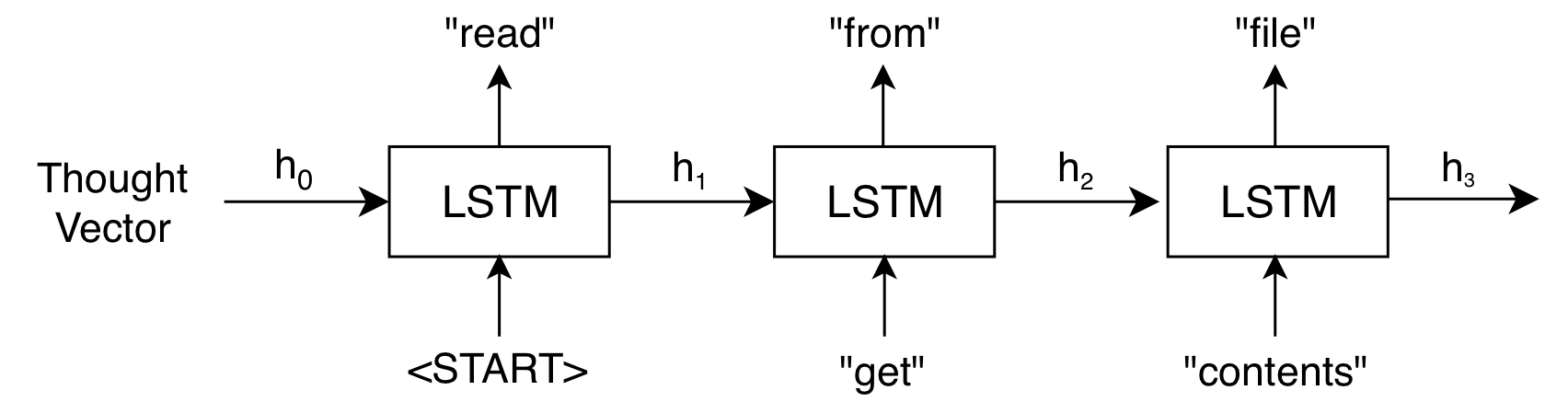}
    \caption{Decoder, in training mode, predicting the phrase ``read from file'' when the correct comment begins with ``get contents.'' Teacher-forcing tends to improve model stability by preventing a sequence from drifting too far off course.}
    \label{fig:decoder_architecture1}
\end{figure}

In training the model, we employ ``teacher-forcing,'' meaning that the LSTM receives the true prior word as input \cite{williams1989learning}. During testing, the LSTM receives the predicted prior word as input. The decoder architecture during training is depicted in Figure~\ref{fig:decoder_architecture1}. For simplicity, we show a single-layer LSTM.

During testing, we conduct a beam search over the output space, per Iyer et al. \cite{iyer2016summarizing}. At each step, we explore the $N$ most-probable next predictions, given the input code and all prior predictions. A beam size of $N$=2 is used.

\subsubsection{\uppercase{Vocabulary Selection}}
\label{sec:vocab}

The model vocabulary is initialized to include every punctuation mark and lowercase English letter.\footnote{All comments are converted to lower case.} At this point, any comment can theoretically be formed by combining elements in the vocabulary. However, learning to combine those elements sufficiently well to predict any word in any comment is a very complex task. We therefore add to the vocabulary the most frequent words and subwords that appear in the training data.

In order to identify the most frequent words and subwords, we first create a word-count dictionary from the comments in the training set. Note that, because of our domain, the initial word-count dictionary is likely to include non-English words such as ``gui''. We then replace elements in that dictionary as follows:
\begin{itemize}
    \item Attempt to split each element into multiple English words.\footnote{We use the US English dictionary from PyEnchant.} E.g., ``FileReader'' is mapped to ``file'' and ``reader''.
    \item If an element ends in ``ing'' or ``ly'', split it into the root word and the suffix. E.g., ``returning'' is mapped to ``return'' and ``ing''.
    \item Attempt to split each element into other elements in the dictionary.\footnote{If an element ends in ``s'' or ``d'' and the remainder of the element appears in the dictionary, it is split accordingly. E.g., ``returns'' is mapped to ``return'' and ``s''.} E.g., ``guiFrame'' is mapped to ``gui'' and ``frame''. These other elements may be non-English terms, which originate from the initial word-count dictionary.
\end{itemize}
\noindent As elements in the dictionary are split, their count values are added to those of their component words or subwords. The elements in the final dictionary with the highest counts are added to our model vocabulary. The number of elements in the vocabulary is determined by validation set performance. 

This method places high-occurrence words in the vocabulary, ensuring that the model learns these words' meanings in a substantive manner. Similarly, the model learns the function of character strings that modify words, such as ``ing'' and ``ly''. Most significantly, our vocabulary creation method allows the model to learn the meaning of technical jargon that is often used to artificially create words. For example, ``gui'' often appears both by itself and in terms such as ``guiFrame''. Models with word-based vocabularies would have to learn the meanings of ``gui'', ``frame'', and ``guiFrame'' independently. Models employing character or n-gram-based vocabularies might struggle with combining ``gui'' and ``frame''. These models will likely have learned that the bigram connecting these two terms (``if'') is primarily used as a separate word, rather than as a character combination linking the two terms.


\section{\uppercase{Experiments}}
\label{sec:experiments}

\noindent We conduct two experiments to evaluate the effectiveness of our model. The following subsections describe the data, present the evaluation mechanism, and note technical details.

	\subsection{Data}

    \noindent We employ two data sets in evaluating our model. First, we use a data set of about 600k Java method/comment pairs that was made available in conjunction with a previous paper on natural language code summarization \cite{hu2018deep}. Second, we use a multilanguage data set of code/comment pairs extracted from many different open-source repositories.

    \subsubsection{Hu et al. Data Set (Java)}
    
    \noindent Previous authors collected a large corpus of matching Java methods and comments from 9,714 GitHub projects \cite{hu2018deep}. They extract a total of 588,108 method/comment combinations and split them 80\%-10\%-10\% into training, validation, and test sets. 
    
    We utilize the same splits to train, validate, and test our model. However, we note that, in the data provided by the authors, some method/comment combinations appear both in the training set and test set. More than 20\% of the examples in the test set (13,000 method/comment pairs) can also be found in the training set. While this phenomenon is likely a reflection of developers' tendency to copy/paste code, it will probably cause overly optimistic results and may privilege models that tend to overfit to their training data.

	\subsubsection{MUSE Corpus Data Set (Java, C++, Python)}
	\label{sec:musecorpus}

    \noindent Separately, we create a large database of code/comment pairs from the MUSE Corpus\footnote{http://muse-portal.net/}, a collection of open-source projects from sites such as GitHub\footnote{https://github.com/} and SourceForge\footnote{https://sourceforge.net/}. We use Doxygen to identify code/comment pairs in files with the extensions ``.java", ``.cpp", ``.cc", and ``.py'', which are the most common extensions for source code files written in the Java, C++, and Python languages. Typically the code is a class, method, or function; however, it can be something as small as a variable declaration. We find 17.4 million code/comment pairs in total: 14.7 million Java, 1.0 million C++, and 1.3 million Python. 

    During the extraction process, we deduplicate the code/comment pairs, removing exact matches. It is common for developers to copy code, comments, or entire source code files, both within and between projects. Lopes et al. find that only 31\% of Java files, 13\% of C/C++ files, and 21\% of Python files on GitHub are non-duplicates \cite{lopes2017dejavu}. The deduplication step during data collection prevents our model from overfitting to the most-copied code and avoids artificially inflating estimates of model accuracy.

    In reviewing the data, we find that, in multisentence comments, the first sentence is usually a description of the code's function and the other sentences contain information that is only tangentially related to the code itself. E.g., ``Returns the version of the file. The only currently supported version is 1000 (which represents version 1).'' Therefore, we elect to use only the first sentence in any extracted comment. We exclude all comments that contain phrases such as ``created by'' and ``thanks to,'' or words such as ``bug'' and ``fix,'' as these do not often describe the code to which they are nominally related. These details are described further in the appendix. Such comments account for about 2\% of Java and C++ samples and 4.5\% of Python samples. 
    
    For similar reasons, we filter out very long (more than 50 tokens) and very short (fewer than 3 tokens) comments. The former tend to contain excessive detail regarding a function's implementation and the latter are often insufficiently specific descriptions (e.g., ``constructor'') or notes to the author (e.g., ``Document me!''). Less than 1\% of samples in the data set have more than 50 tokens and about 10\% have fewer than 3 tokens. This holds for the sets of Java, C++, and Python samples, individually, as well.

    We also examine the code samples in the data set. We find that short code samples usually contain only single decontextualized words, e.g., ``ITEM''. Therefore, we exclude samples where the code has fewer than 8 characters. Less than 0.2\% of the samples in the data set have code with fewer than 8 characters. Similarly, we find that very long code samples are unlikely to be well-summarized by a single comment, so we exclude samples where the code contains more than 4,096 characters. Approximately 7\% of Java samples, 2.5\% of C++ samples, and 5.5\% of Python samples are excluded on this basis.
    
    Table~\ref{tab:a} shows the number of code/comment pairs in our data set, by language. We utilize about 85\% of the collected samples of each language; the other 15\% are excluded by the filters described previously.

    \begin{table}[!htb]
    \caption{Number of observations collected from MUSE Corpus and number of observations used to train, validate, and test the model.}\label{tab:a}
        \begin{center}
          \begin{tabular}{ |l|r|r| } 
           \hline
           \multicolumn{1}{|p{1.75cm}|}{\raggedright \textbf{Language}}
           & \multicolumn{1}{|p{1.75cm}|}{\raggedleft \textbf{Collected}}
           & \multicolumn{1}{|p{1.75cm}|}{\raggedleft \textbf{Used}}
           \\\hline
           C++ & 1,050,077 & 918,583 \\
           \hline
           Java & 14,709,616 & 12,461,021 \\
           \hline
           Python & 1,325,845 & 1,121,421 \\
           \hline
          \end{tabular}
        \end{center}
    \end{table}

\begin{figure}[h]
   \centering
   \includegraphics[scale=.60]{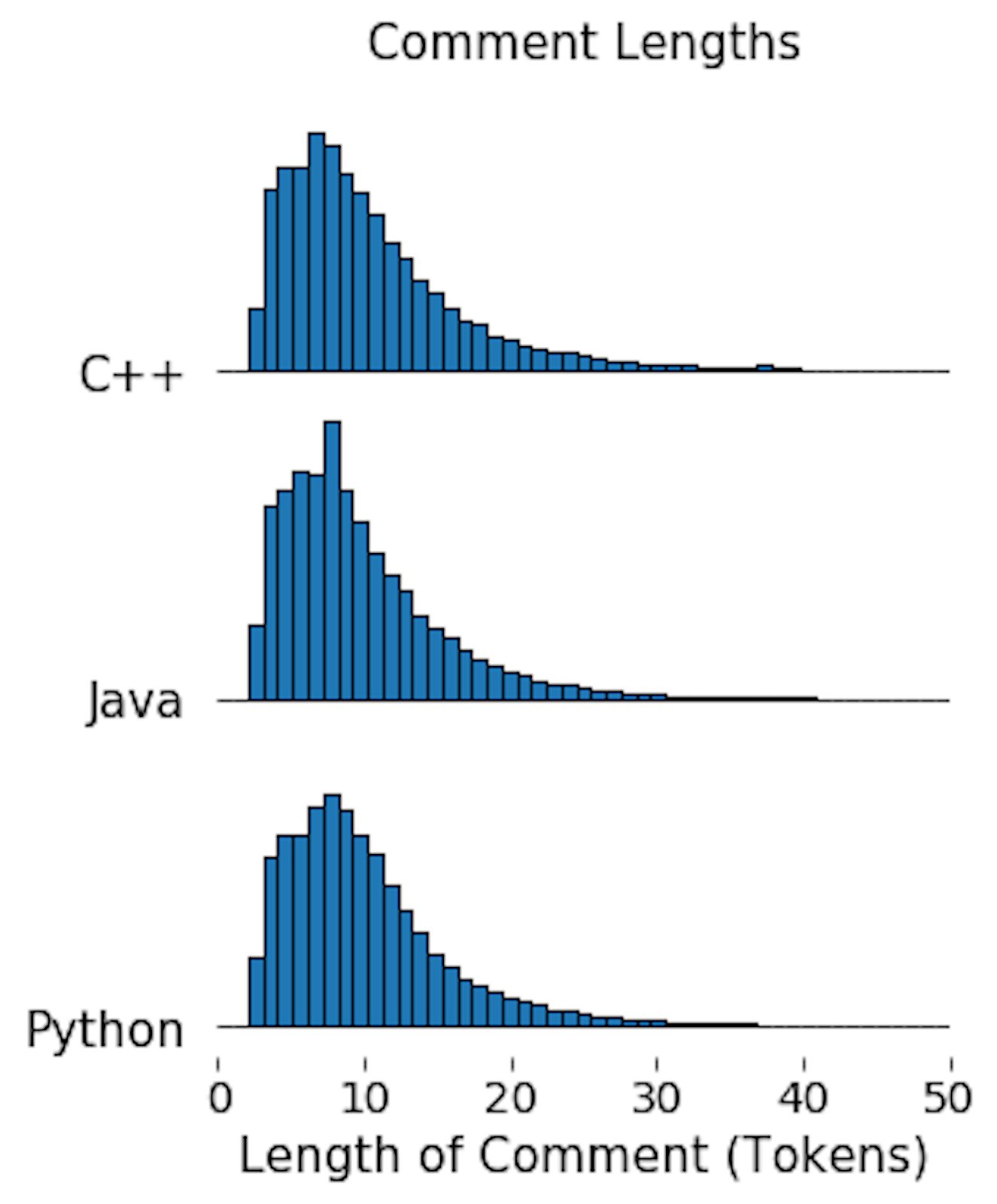}
    \caption{Distribution of comment lengths in the filtered data set. Most comments contain fewer than 20 tokens.}
    \label{fig:comment_lengths}
\end{figure}

\begin{figure}[h]
   \centering
   \includegraphics[scale=.60]{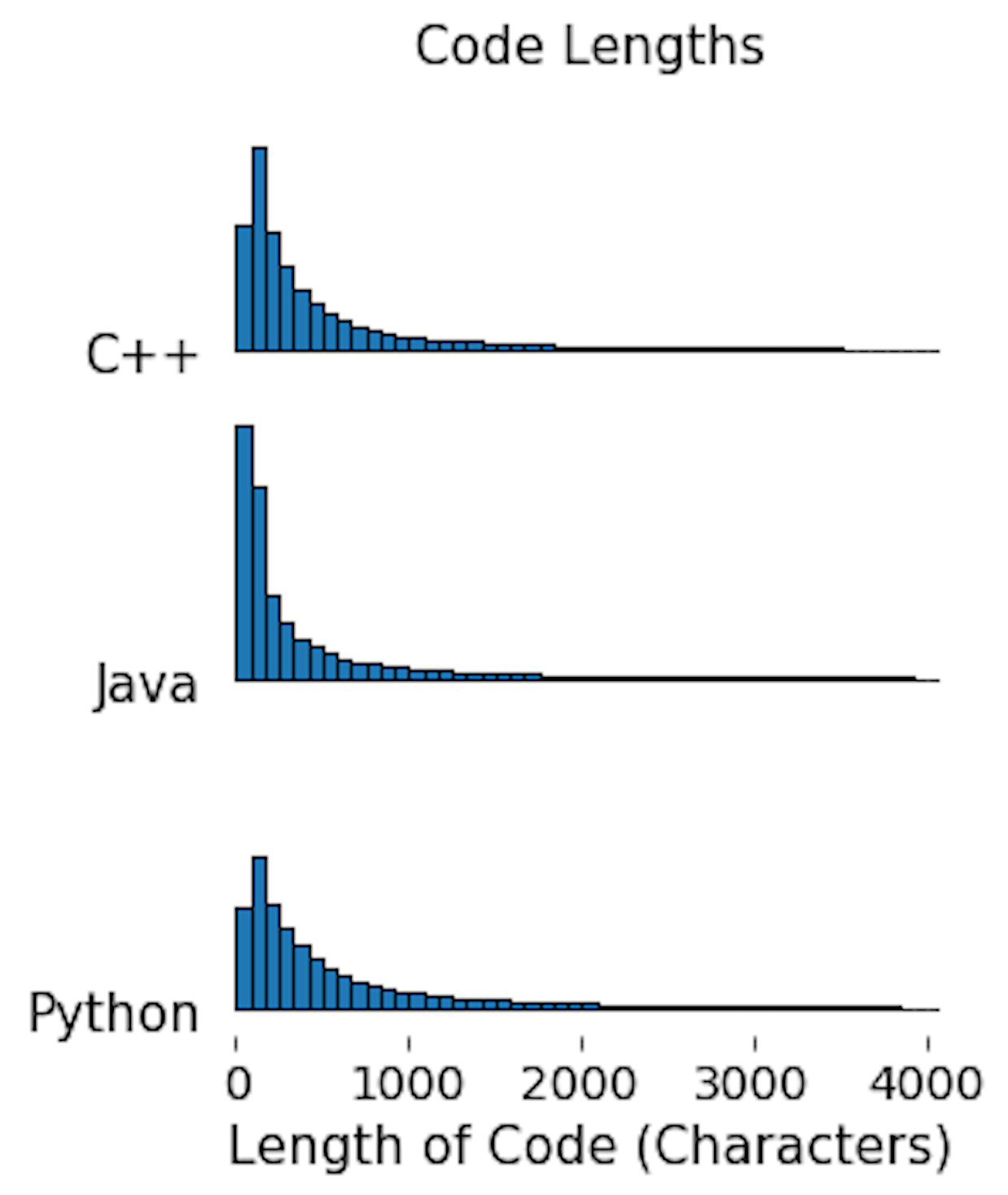}
    \caption{Distribution of code lengths in the filtered data set. Most code examples contain fewer than 1,000 characters.}
    \label{fig:code_lengths}
\end{figure}

     Figure~\ref{fig:comment_lengths} shows the distribution of comment lengths after our filtering. It is highly skewed; the vast majority of comments contain fewer than 20 tokens. The length of code examples, shown in Figure~\ref{fig:code_lengths}, is similarly skewed. Nearly all code examples contain fewer than 1,000 characters.

	\subsection{Evaluation}
	
	To evaluate the effectiveness of our method, we calculate the bilingual evaluation understudy (BLEU) score for each test example. This metric is often used to evaluate machine translation results \cite{papineni2002bleu,callison2006re}. Recent work in source code summarization has used BLEU scores or close variants as the primary evaluation metric \cite{hu2018deep,iyer2016summarizing}. 
	
	A BLEU score compares a predicted sentence to a reference sentence (both tokenized). The score is based on the average n-gram precision of the predicted sentence as compared to the reference sentence. This provides high scores to predictions in which most n-grams are contained in the reference. The BLEU score calculation also incorporates a brevity penalty that penalizes overly short predictions. Thus, the two components balance each other -- one rewarding the model for predicting \textit{only} n-grams that are contained in the reference sentence and the other rewarding the model for making appropriately lengthy predictions.
	
	We use BLEU-4, meaning the BLEU score calculation includes 4-grams and all smaller n-grams. Consistent with Hu et al., we do not use smoothing to resolve the lack of higher order n-gram overlap. Mathematically, we calculate BLEU-4 as,
	\begin{equation}
	BLEU = B e^{\sum_{n=1}^4 w_n \log(p_n)}
    \end{equation}

    \begin{equation}
    B=\left\{
      \begin{array}{@{}ll@{}}
        1, & \text{if}\ c > r \\
        e^{1-r/c}, & \text{otherwise}
      \end{array}\right.
    \end{equation} 
    
    \noindent where $p_n$ is the proportion of n-grams in the prediction that are also in the reference, $w_n$ is the weight associated with those n-grams, c is the length of the prediction, and r is the length of the reference \cite{hu2018deep,papineni2002bleu}.
	
	\subsection{Modeling and Training Details}
	\label{sec:mandtdetails}
    
    We select hyperparameters based on model performance on held-out validation sets. On that basis, we embed input characters in a 16-dimensional space. We also tested 8 and 32 dimensions. We choose a thought vector and LSTM decoder of size 1024. We also tested vectors of sizes 512 and 256. For the Hu et al. data set, the decoder has a single layer of LSTM units; for the MUSE Corpus data set, the decoder consists of two layers of LSTM units. In both instances, we tested models with one, two, and three such layers.
        
    We similarly choose the size of the vocabulary based on validation set performance. For the Hu et al. data set, the vocabulary includes all words and subwords that appear at least 10 times in the training set. For the MUSE Corpus data set, the vocabulary includes all words and subwords that appear at least 500 times in the training set. The number of appearances of a word or subword is calculated using the method described in Section \ref{sec:vocab}.
        
    For the Hu et al. data set, the model is trained for 25 epochs (passes through the entire training set). After each epoch, the model is validated using the entire validation set. For the MUSE Corpus data set, the model is trained for 100 rounds, where each round of training is composed of 100,000 examples. After each round, the model is validated on 9,600 examples. For both data sets, the best-validated model (based on cross-entropy loss) is selected for use in testing. Batches of size 32 are used throughout.

\section{\uppercase{Results}}

\subsection{Quantitative Results}

    \noindent We report test set BLEU scores and comment entropy in Tables \ref{tab:huresults} and \ref{tab:museresults} for the Hu et al. data set and the MUSE Corpus data set, respectively. Both BLEU scores and comment entropy are based on a tokenization (into punctuation and space-separated words) of the actual and predicted comments. The comment entropy (in bits) is a measure of the amount of information contained in the comments of each test set.
    
    For the Hu et al. data set, we train, validate, and test our model using the provided subsets of the data. Our method achieves parity with Hu et al.'s model, while avoiding the language-specific code parsing and AST creation that their model requires.
    
    \begin{table}[!htb]
    \caption{Results on Hu et al. data set.}\label{tab:huresults}
        \begin{center}
          \begin{tabular}{ |l|r|r|r| } 
           \hline
           \multicolumn{1}{|p{1.5cm}|}{\raggedright \textbf{ } \\ \textbf{Language}}
           & \multicolumn{1}{|p{1.3cm}|}{\raggedleft \textbf{ } \\ \textbf{BLEU}}
           & \multicolumn{1}{|p{1.3cm}|}{\raggedleft \textbf{Hu et al.} \\ \textbf{BLEU}}
           & \multicolumn{1}{|p{1.3cm}|}{\raggedleft  \textbf{ } \\ \textbf{Entropy}}
           \\\hline
           Java & 38.63 & 38.17 & 104.92 \\
           \hline
          \end{tabular}
        \end{center}
    \end{table}
    
    For the MUSE Corpus data set, we randomly select 50,000 observations per language to serve as test sets. We then split the remaining data 80\%-20\% into training and validation sets.  
    
    The model clearly performs better on code written in Java and C++ than it does on code written in Python. We hypothesize that this is due to the strong syntactic relationship between the former two languages. The model has seen twice as many C++ and Java code/comment pairs as Python code/comment pairs and thus is better trained to summarize code with a C++/Java-like style and syntax. These results suggest that, if we want our model to perform well on additional languages, we will need to collect an appropriate quantity of training data. However, we may be able to leverage related languages, if code from a particular language is scarce.
    
    \begin{table}[!htb]
    \caption{Results on MUSE Corpus data set.}\label{tab:museresults}
        \begin{center}
          \begin{tabular}{ |l|r|r| } 
           \hline
           \multicolumn{1}{|p{1.5cm}|}{\raggedright \textbf{Language}}
           & \multicolumn{1}{|p{1.5cm}|}{\raggedleft \textbf{BLEU}}
           & \multicolumn{1}{|p{1.5cm}|}{\raggedleft \textbf{Entropy}}
           \\\hline
           C++ & 40.34 & 100.18 \\
           \hline
           Java & 41.13 & 93.46 \\
           \hline
           Python & 36.69 & 102.27 \\
           \hline
          \end{tabular}
        \end{center}
    \end{table}
    
    Additionally, we analyze how our model's performance relates to the amount of information contained in each language's comments. We calculate the entropy of a programming language $k$'s comments as,
    
    \begin{equation}
    E_k = w_k  \sum_{i=0}^V p_{ik} log(p_{ik})
    \end{equation} 
    
    \noindent where, for comments associated with code written in language $k$, $w_k$ is the average length of a comment and $p_{ik}$ is the ratio of the number of occurrences of token $i$ to the total number of tokens in all comments. Within the MUSE Corpus, there is an inverse relationship between the BLEU scores that our model obtains for comments of a given language and the entropy of that programming language's comments. This is consistent with our expectations; the more information content comments have, the harder they are to predict.

\subsection{Qualitative Results}
    
    While BLEU scores are one of the most common machine translation evaluation metrics, they sometimes fail to fully capture the accuracy of a predicted comment \cite{callison2006re}. BLEU scores do not account for synonyms or local paraphrasing. However, using human reviewers with a specialized skill set (programming) is prohibitively time-consuming and expensive. In an effort to demonstrate the quality of our results, we present a group of randomly selected examples from each language in Table \ref{tab:museexampleresults1} (Java), Table \ref{tab:museexampleresults2} (C++), and Table \ref{tab:museexampleresults3} (Python).
    
   Model-generated comments appear to fall into roughly three categories: (1) correct, (2) related, and (3) incorrect/uninformative. The first several examples in each table are of ``correct'' comments -- those that capture the intent of the original comment. As might be expected, correct comments are often associated with code that has a specific and commonly used functionality, e.g., creating a file. The next several examples in each table are of ``related" comments -- those that are thematically related to the original comment, but do not capture its full meaning. In these cases, the model likely identified a few terms in the code that suggested the correct topic, but failed to comprehend the fuller context. The last few examples in each table are of ``incorrect/uninformative" comments -- those that are either fully unrelated to the original comment or are very poorly written. Predicted comments that are unrelated to the original seem to often be produced when the model is asked to summarize relatively unusual code. In these cases, it appears to generate a comment regarding the most-similar ``common'' concept.
    
    As is obvious from a few of the examples in each table, the model sometimes produces results that do not constitute proper English sentences. While there is no a priori reason that it should only produce grammatically correct predictions, one might expect it to have learned proper English construction from the many example comments on which it was trained. Unfortunately, many comments that appear in the training data include grammatical errors, typos, etc. and the model has learned to mimic these. This is especially true for comments associated with Python code. Additionally, for a given piece of code, the model's primary goal is to generate a comment that minimizes expected cross-entropy loss when compared to the true comment. The expected-loss-minimizing comment will not necessarily have proper English construction. For example, the model sometimes repeats particular words; it has confidence that particular word will be in the output, but is not sure where in the output it belongs. The model tries to maximize the probability of generating the correct output by putting the word everywhere. This and other behaviors are likely artifacts of cross-entropy loss, which penalizes errors at the word-level, rather than at the level of sentence-meaning, leading to unusual word combinations. We hypothesize that a loss function that accounts for predictions at all steps simultaneously would greatly improve the model's output; we leave this as future work.
    
    \begin{table*}[!htb]
      \caption{Selected examples from MUSE Corpus test set for code written in Java.}\label{tab:museexampleresults1}
      \begin{center}
          \begin{tabular}{ |p{6.25cm}|p{6.25cm}|p{1.75cm}| } 
           \hline
           \textbf{Actual Comment} & \textbf{Predicted Comment} & \textbf{Assessment} \\
           \hline
           change the length of the file . & sets the length of the file . & correct \\
           \hline
            convenience method to create a file & creates a file . & correct \\
           \hline
           custom layout for an item representing a bookmark in the browser . & adds a bookmark to the list . & related \\
           \hline
           calculate the distance in metres to the rhumb line defined by coords a and b . &
           returns the distance between two points . & related \\
           \hline
           returns a stream of random sample of all public statuses . & returns the currently active stream . & related \\
           \hline
           serializer adapter on top of spring ' s o / x mapping . & provides serialization methods for serializing objects . & related \\
           \hline 
           calls the specified method with the given params .  & returns a string containing the string representation of the string . & incorrect \\
           \hline
           returns the xml sensordatatypeindex for all sensor data . & returns the string used for creating objects . & incorrect \\
           \hline
           compute the optimized gradient & returns the gradient gradient gradient gradient gradient gradient gradient . & incorrect \\
           \hline
          \end{tabular}
        \end{center}
    \end{table*}
    
    \begin{table*}[!htb]
      \caption{Selected examples from MUSE Corpus test set for code written in C++.}\label{tab:museexampleresults2}
      \begin{center}
          \begin{tabular}{ |p{6.25cm}|p{6.25cm}|p{1.75cm}| }   
           \hline
           \textbf{Actual Comment} & \textbf{Predicted Comment} &
           \textbf{Assessment} \\
           \hline
           set the maximal width of a tab . & sets the width of the tab . & correct \\
           \hline
           resumes the thread . & starts a new thread . & related \\
           \hline
            get a chunk of memory from free list cache . & returns the number of bytes that can be used for writing . & related \\
           \hline
           construct an instruction text to the ui . & creates a new text object . & related \\
           \hline
           insert a row cut unless it is a duplicate - cut may get sorted . & creates a new matrix . & related \\
           \hline
           create a tileset from a pixmap . & creates a mapping from a bitmap to a pixmap . & related \\
           \hline
        construct a cvc ado request from a der encoded cvc ado request file . & creates a new instance . & incorrect \\
            \hline
        serialize the pbmap .  & updates the mapping of the mapper to the given string . & incorrect \\
           \hline
           convert a qstringlist to a vector of string & returns a list of all the names of the names of the names of the names of the names of the names of the names of & incorrect \\
           \hline
          \end{tabular}
        \end{center}
    \end{table*}
    
    \begin{table*}[!htb]
      \caption{Selected examples from MUSE Corpus test set for code written in Python.}\label{tab:museexampleresults3}
      \begin{center}
          \begin{tabular}{ |p{6.25cm}|p{6.25cm}|p{1.75cm}| } 
           \hline
           \textbf{Actual Comment} & \textbf{Predicted Comment} &
           \textbf{Assessment} \\
           \hline
           raises an assertion error if two items are equal . & asserts that the values of the values are equal . & correct \\
           \hline
           get the identifier assigned to the message .  & returns the id of the underlying dbm . & correct \\
           \hline
           print elapsed , user , system , and user + system times & prints a list of timestamps . & related \\
           \hline
           return the record number based on a record number or name . & returns the record type for the given record . & related \\
           \hline
           abstraction for a collection of reactions . & returns a list of action actions . & related \\
           \hline
           display an information dialog box . & returns a string containing the contents of the given string . & incorrect \\
           \hline
           create the property instance :  & the meta object literal for the ' & incorrect \\
           \hline
           enables / disables the item . & enable / disable enabled enabled & incorrect \\
           \hline
           return entry point group for given object type . & returns a group of groups for the given group . & incorrect \\
           \hline
          \end{tabular}
        \end{center}
    \end{table*}
    
\section{\uppercase{Related Work}}

\subsection{Source Code Summarization}

\noindent A wide variety of techniques have been applied to create natural language summaries of source code. Some of the first research in the area relied heavily on intermediate representations of code and template-based text generation. For example, Sridhara et al. utilize the Software Word Usage Model, which ingests ASTs, control flow graphs, and use-define chains, to generate human-readable summaries of Java methods \cite{sridhara2010towards}. This technique works reasonably well, according to the developers engaged to review its output, but it is specifically engineered for Java method summarization and its descriptiveness is potentially inhibited by its output templates. By contrast, our method generalizes to any programming language and is capable of generating arbitrarily complex natural language descriptions.

Techniques primarily derived from the information retrieval and natural language processing domains have also been applied to the task \cite{allamanis2017survey,allamanis2015suggesting,movshovitz2013natural,hindle2012naturalness}. Haiduc et al. and De Lucia et al., for example, create extractive summaries based on the position of terms in the source code and the results of common information retrieval techniques \cite{de2012using,haiduc2010supporting,haiduc2010use}. Movshovitz-Attias and Cohen apply latent Dirichlet allocation topic models to aid in comment completion \cite{movshovitz2013natural}. Allamanis et al. develop a language model to suggest method and class names \cite{allamanis2015suggesting}. These works target similar, but more well-constrained tasks than the one that we have addressed. None of them allow arbitrary code to be used as input or produce open-ended natural language output.


More recently, researchers have cast source code summarization as a machine translation problem and applied deep learning models reminiscent of those used to translate between two natural languages. Our method is most in-line with these approaches. Iyer et al. develop CODE-NN, an LSTM-based neural network \cite{iyer2016summarizing}. CODE-NN's attention mechanism allows it to focus on different tokens in the source code while predicting each word in the summary. We do not offer comparisons of our model to CODE-NN using Iyer et al.'s data sets because of their relatively limited size. Hu et al. propose several LSTM-based encoder/decoder models \cite{hu2018deep}. Hu's most effective model, DeepCom, uses a novel AST traversal method to represent ASTs as sequences, which are then used as input to a recurrent neural network. 
However, neither CODE-NN nor any of the models proposed by Hu et al. achieve language agnosticism, because they require language-specific intermediate representations of the input code. We bypass this limitation by utilizing a source-code-as-text approach that avoids not only language-specificity, but all syntactic constraints. 

\subsection{Open-Vocabulary Machine Translation}

Our model's decoder draws on previous research into 
vocabulary selection and modeling in machine translation systems. It is conventional for closed-vocabulary models to utilize word-based vocabularies, i.e., every element in the vocabulary is a single word \cite{tu2017neural,serban2017hierarchical}. Most natural language data sets, however, contain far too many unique words for all of them to be included in the vocabulary without radically increasing model complexity. Usually, the words that occur most frequently in the training data will be included in the vocabulary; all other words will be mapped to an ``unknown word'' token \cite{jean2015montreal,luong2014addressing}. This is particularly problematic in the context of source code summarization, because the distribution of words in source code comments has a very long tail (i.e., a lot of very infrequently used terms). Mapping all of these terms to the ``unknown word'' token would severely inhibit the model's ability to accurately describe a piece of code. It is especially problematic because infrequently used words are often indicative of a code's context and thus convey the most information.

Open-vocabulary models like ours try to avoid unknown word tokens, seeking the ability to predict any word, even those not observed during training. Some open-vocabulary models generate character-by-character (instead of word-by-word) output, so that any arbitrary character sequence could potentially be produced \cite{matthews2016synthesizing,ling2015finding,ling2015character}. Other systems employ a vocabulary composed of character n-grams that can be combined to form words \cite{wu2016google,sennrich2015neural}. 
Both of these strategies, however, prevent models from learning meaning at the word level, hindering their understanding of complete words and compounds formed from multiple words. The latter occur frequently in source code comments. Because our method combines words, subwords, and characters, the model is able to understand the meaning of individual words, how to utilize subwords appropriately, and how to spell words (in part or in whole) when necessary. 

\section{\uppercase{Conclusions and Future Work}}

We present a novel encoder/decoder model capable of summarizing arbitrary source code. We demonstrate results comparable to the state-of-the-art for a single-language (Java), while avoiding the cumbersome parsing required by previous source code summarization models. Additionally, we present the first results on a data set containing multiple programming languages. The model's effectiveness under those conditions demonstrates its ability to learn the function of a piece of source code, regardless of the code's syntax and style. Finally, while previous large-scale work has focused on Java, we provide the first baselines for summarization of C++ and Python code.

Our methods constitute the first success in performing language-agnostic source code summarization; however, there are a number of changes to the model architecture that may enhance performance. As discussed earlier, we could modify the loss function to operate on all steps of a prediction simultaneously, rather than on a single token at a time. This change would likely improve both linguistic correctness and model accuracy, because the loss function would better reflect the way that comments (and natural language statements more generally) are actually written. We could also incorporate a ``copying" mechanism, per Gu et al., to enable the summary to directly reference terms in the source code \cite{gu2016incorporating}. This mechanism is likely to help the model handle very specific and rare code more effectively, because the model would be able to copy unusual words, such as proper names, directly from the source code.

Although there are certainly ways to improve the model, given its initial success, we believe that there is significant room for it to be applied to ongoing software engineering work. For example, source code summarization might prove useful as a plug-in for an integrated development environment or as a component of a version control system. In a deployed system, one could potentially make use of templates to constrain model output to a specific desired form (as was done in early summarization work), thereby lowering the complexity of the prediction task. A model of the same style could also be used to generate summaries of source code at the document or project level. Such higher-level summaries might speed up the process of onboarding new developers and generally enable easier navigation of code bases. Finally, one might even utilize this type of model to perform translation in the other direction, between natural language and, for example, pseudocode. 

\section*{ACKNOWLEDGEMENTS}

\noindent This project was sponsored by the Air Force Research Laboratory (AFRL) as part of the DARPA MUSE program.

\bibliographystyle{apalike}
{\small
\bibliography{bibliography}}

\section*{APPENDIX}

\noindent Below, we provide additional technical details on our data processing procedures.

\textbf{Comment Tokenization.} In order to create the word-count dictionary described in Section \ref{sec:vocab}, comments must be tokenized. First, all letters in a comment are converted to lower case. Then, per Hu et al., we tokenize the comment by considering each punctuation mark and space-separated term to be an individual token.

\textbf{MUSE Corpus Sentence-Break Identification.} Per Section \ref{sec:musecorpus}, when there are multiple sentences in an extracted comment, we use only the first sentence in our modeling and evaluation process. We identify the end of the first sentence with any of the following strings:
\begin{itemize}[noitemsep]
    \item ``.''
    \item ``\textbackslash n \textbackslash n''
    \item ``:param''
    \item ``@param''
    \item ``@return''
    \item ``@rtype''
\end{itemize}

\textbf{MUSE Corpus Comment Filtering.} Per Section \ref{sec:musecorpus}, we exclude code/comment pairs if the comment contains specific words or phrases. In particular, we exclude any code/comment pair in which the comment contains any of the following strings:
\begin{itemize}[noitemsep]
    \item ``created by"
    \item ``thanks to"
    \item ``precondition"
    \item ``copyright"
    \item ``do not remove"
    \item `` bug "
    \item `` fix "
    \item ``?"
    \item ``-$>$"
    \item ``$>>>$"
    \item ``(self,"
\end{itemize}

\vfill
\end{document}